\documentclass{article}

\usepackage[final,nonatbib]{neurips_2024}

\usepackage[utf8]{inputenc} % allow utf-8 input
\usepackage[T1]{fontenc}    % use 8-bit T1 fonts
\usepackage{fontawesome}
\usepackage{hyperref}       % hyperlinks
\usepackage{url}            % simple URL typesetting
\usepackage{booktabs}       % professional-quality tables
\usepackage{amsfonts}       % blackboard math symbols
\usepackage{nicefrac}       % compact symbols for 1/2, etc.
\usepackage{microtype}      % microtypography
\usepackage{xcolor}
\usepackage{hyperref}
\hypersetup{
    colorlinks=true,      % Enable colored links
    linkcolor=blue,       % Color for internal links (like table of contents)
    citecolor=blue,       % Color for citations
    urlcolor=blue         % Color for URLs
}
\usepackage{color,colortbl,soul}
\definecolor{Gray}{gray}{0.9}
\usepackage{graphicx}
\usepackage{tabularx}
\usepackage{multirow}
\usepackage{amsmath}
\usepackage{subcaption}
\usepackage{tikz}
\usetikzlibrary{positioning}
\usepackage[numbers]{natbib}

\title{A Benchmark for Long-Form Medical Question Answering}

\author{%
  Pedram Hosseini\thanks{Corresponding author: \href{mailto:pedram@lavita.ai}{\texttt{pedram@lavita.ai}}} \\
  Lavita AI \\
  \And
  Jessica M. Sin \\
  Dartmouth Hitchcock Medical Center \\
  \AND
  Bing Ren \\
  Dartmouth Hitchcock Medical Center \\
  \And
  Bryceton G. Thomas \\
  Dartmouth Hitchcock Medical Center \\
  \And
  Elnaz Nouri \\
  Lavita AI \\
  \And
  Ali Farahanchi \\
  Lavita AI \\
  \And
  Saeed Hassanpour \\
  Dartmouth College \\
}

\begin{document}

\maketitle

\begin{center}
    \faGithub~Code \& Data:~\url{https://github.com/lavita-ai/medical-eval-sphere}
\end{center}

\begin{abstract}
There is a lack of benchmarks for evaluating large language models (LLMs) in long-form medical question answering (QA). Most existing medical QA evaluation benchmarks focus on automatic metrics and multiple-choice questions. While valuable, these benchmarks fail to fully capture or assess the complexities of real-world clinical applications where LLMs are being deployed. Furthermore, existing studies on evaluating long-form answer generation in medical QA are primarily closed-source, lacking access to human medical expert annotations, which makes it difficult to reproduce results and enhance existing baselines. In this work, we introduce a new publicly available benchmark featuring real-world consumer medical questions with long-form answer evaluations annotated by medical doctors. We performed pairwise comparisons of responses from various open and closed-source medical and general-purpose LLMs based on criteria such as correctness, helpfulness, harmfulness, and bias. Additionally, we performed a comprehensive LLM-as-a-judge analysis to study the alignment between human judgments and LLMs. Our preliminary results highlight the strong potential of open LLMs in medical QA compared to leading closed models.
\end{abstract}

\section{Introduction}
\label{sec:intro}

The majority of existing LLM evaluation benchmarks in medical question answering (QA) have focused on automatic metrics and multiple-choice questions~\cite{manes-etal-2024-k,kim-etal-2024-medexqa,shi-etal-2024-medical,sviridova-etal-2024-casimedicos}. Although valuable, such metrics and question formats fall short of reflecting the realistic settings of real-world clinical scenarios~\cite{xiong2024benchmarking,shi-etal-2024-medical} and do not fully assess or capture the nuances and factual accuracy of LLMs in the medical domain~\cite{kim-etal-2024-medexqa,wang-etal-2024-imapscore,yang-etal-2024-kg}. Additionally, there are concerns about the potential leakage of well-known benchmarks into the training data of LLMs that are subsequently evaluated on those same benchmarks~\cite{deng2023benchmark}. Furthermore, benchmarks that have not leaked may contain label errors or be outdated~\cite{saab2024capabilities}, leading to flawed and unrealistic evaluations. Moreover, the limited research that has explored human evaluation of long-form medical QA has not made the associated labels publicly available, hindering reproducibility and the ability to study human annotations for insights to inform future work.

\begin{figure}[]
    \vspace*{-\baselineskip}
    \centering
    \includegraphics[width=0.85\linewidth]{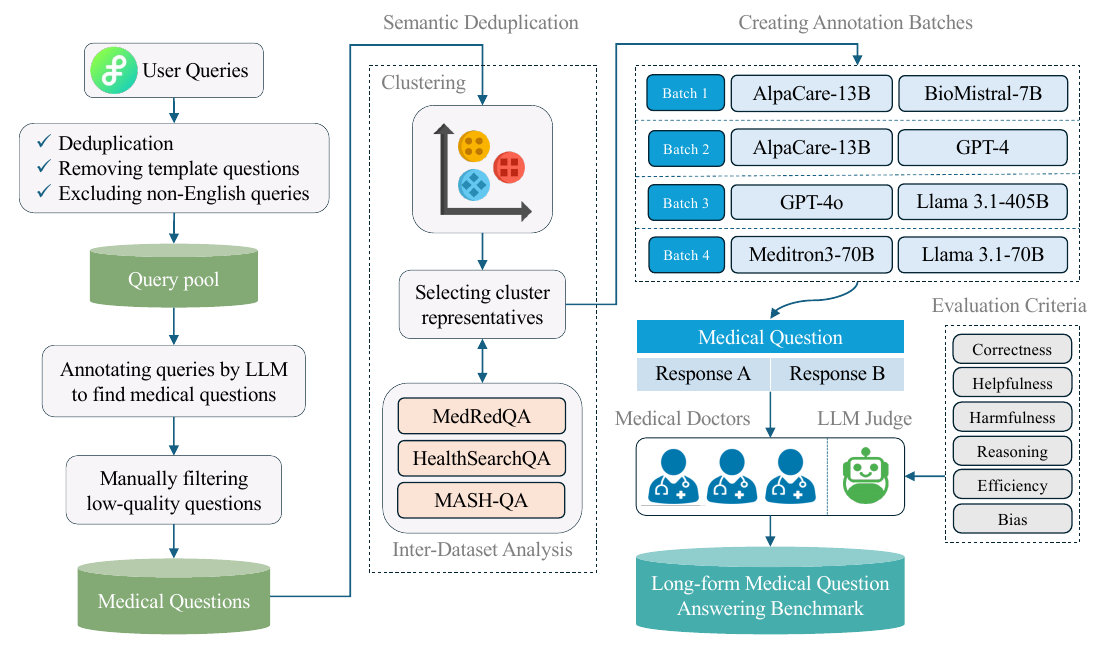}
    \caption{Overview of our benchmark creation process}
    \label{fig:benchmark-overview}
\end{figure}

To address these challenges, we introduce a new publicly available benchmark of real-world consumer medical questions with long-form answer evaluation, annotated by medical doctors. An overview of the process of building our benchmark is shown in Figure~\ref{fig:benchmark-overview}. Our contributions are as follows:
\begin{itemize}
\item We develop a publicly available benchmark for long-form medical QA, based on real-world consumer health and medical questions.
\item We release the medical doctor annotations to the research community, along with our growing collection of real-world consumer medical questions. % We additionally share direct feedback and lessons from our medical doctors for future evaluation improvements.
\item We conduct a comprehensive analysis comparing human experts and LLM-as-a-judge for evaluating long-form answer generation in medical questions.
\end{itemize}

\section{Data Preparation}
\begin{figure}[h]
    \centering
    \includegraphics[scale=0.5]{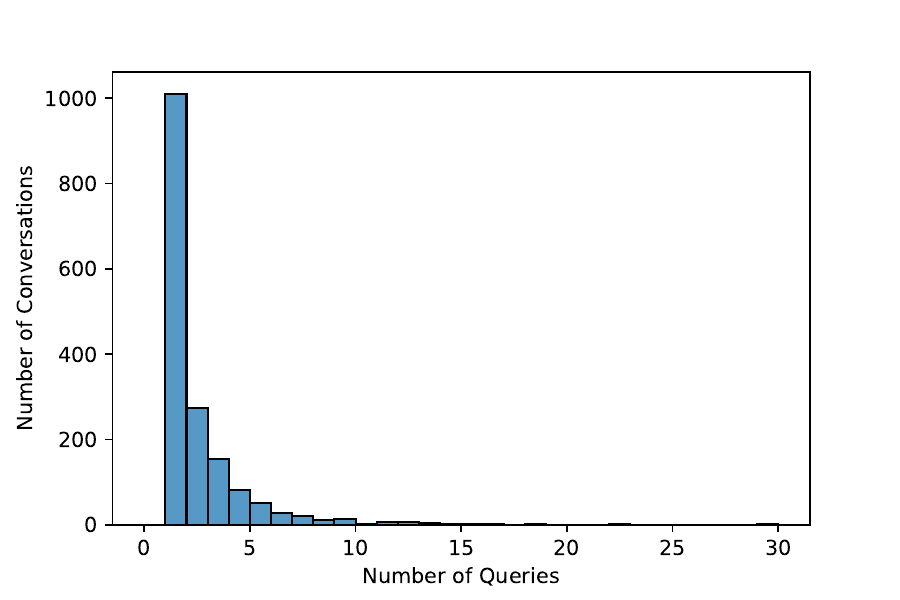}
    \caption{The distribution of queries in conversations (excluding outliers)}
    \label{fig:query_distribution}
\end{figure}

To create a dataset of real-world consumer medical questions, we collected queries by users on our platform, \emph{Lavita Medical AI Assist}.\footnote{\url{https://assist.lavita.ai/}} In this phase, we collected all queries from 2023-10-31 (the date we launched our platform) to 2024-02-12 containing 4,271 inputs in 1,693 conversations. Conversations could be single-turn (1,011) or multi-turn (682). However, for now, we ignored the conversation types and processed queries independently, without considering whether they were part of the same conversation. The distribution of queries in conversations is shown in Figure~\ref{fig:query_distribution}. After collecting the queries, we deduplicated them, removed those from our sample question pool,\footnote{By default, we show a few randomly selected questions on our website from a pool of trending medical questions. If a user's question is from the pool, we remove it from our dataset.} and filtered non-English entries using Lingua.\footnote{\url{https://github.com/pemistahl/lingua-py}} This step resulted in 2,698 queries.

\subsection{Finding Medical Questions}
User queries could be quite noisy and not all of them ask a clear and direct medical or health-related question. Additionally, some queries, even though medical, contain grammatical and/or spelling errors. Since manually filtering and correcting these cases is a time-consuming process, we prompted GPT-4\footnote{\texttt{gpt-4-0125-preview}} to 1) tell if a user query contains a direct medical or health-related question, and 2) correct any grammatical or spelling errors in queries by retaining their original information and meaning. The full prompt template is shown in Figure~\ref{fig:query_detection_prompt_template}.

\subsubsection{Human Validation}
Before relying on GPT-4's annotations for medical question detection, we created a random sample of queries and had two human annotators label the same questions we asked GPT-4. We created a representative sample size of 337 based on Cochran`s formula~\cite{cochran1977sampling}. There is a 94\% agreement among the two human annotators on whether a query contains a direct medical and health-related question. And, in those cases the two human annotators agreed, there's also 91\% agreement with GPT-4's predictions. One common pattern we observed in disagreements between humans and GPT-4 was in cases in which it is hard to find a clear, direct, and independent question. For example, \textit{What dose of spermidine did they use?} Even though this question is health-related and medical, it is not clear who \emph{they} is referring to. These cases happen especially when a question is part of a multi-turn conversation which is not the focus of our study here. Considering the high agreement between human annotators and GPT-4, we separated all queries labeled by GPT-4 as a medical question.

\subsubsection{Quality Check}
To ensure the grammatically corrected version of queries by the model are not significantly different than the original ones, we computed\footnote{We used the \textit{SequenceMatcher} from the \textit{difflib} library.} the similarity between the original and corrected queries with an 85\% threshold after manually inspecting some samples to determine a reasonable threshold. This process resulted in 1,446 queries identified as medical questions. It is worth noting that using GPT-4's annotation, we were able to automatically cut 46\% of queries that are non-medical. And, our prompt and pipeline could also be used in the future to filter out non-medical questions at a higher scale. Even though this is not the focus of our study, this experiment also shows the potential of using highly capable LLMs for medical question identification or classification.

To further check the quality of our questions, as the final step, one human annotator went through the 1,446 medical questions and removed any question that might have been wrongfully annotated by GPT-4 as a direct medical question. This process resulted in 1,298 remaining higher-quality medical questions.

\subsection{Semantic Deduplication}
In order to minimize similarity among the questions in our benchmark and to perform semantic deduplication, we conducted an intra-dataset similarity analysis to assess the degree of semantic similarity between the questions. For this analysis, we first computed the embeddings of all questions using OpenAI's \textit{text-embedding-3-large} model with a dimension size of 1024. We then constructed a distance matrix of the embeddings using cosine distance, defined as 1.0 minus the cosine similarity between two embedding vectors. Then we clustered embeddings based on DBSCAN~\cite{ester1996density}\footnote{We used the scikit-learn implementation~\cite{scikit-learn}.} with:
\begin{align*}
\textit{threshold} &= 0.75 \\
\textit{eps} &= 1.0 - \textit{threshold} = 0.25
\end{align*}%with $eps=0.25$
We tried multiple thresholds and manually checked the clusters every time to ensure clusters were neither too specific nor too broad. This process resulted in 1,077 clusters. Then we selected the medoid of each cluster as the cluster's representative. As a result, we ended up with 1,077 more semantically distinct questions.

\subsection{Inter-Dataset Similarity Analysis}
\begin{figure}[]
    \centering
    \includegraphics[scale=0.55]{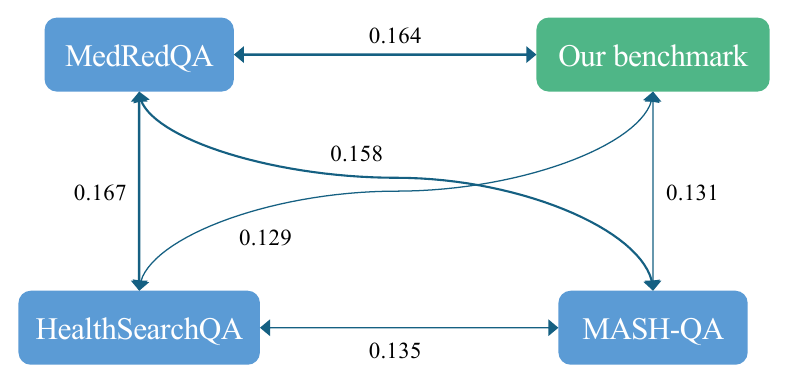}
    \caption{Similarity graph of consumer medical question answering datasets}%Each node represents a dataset and the number on the edge shows the mean similarity score of a pair of datasets}
    \label{fig:similarity-graph}
\end{figure}

We also compared our dataset with three other well-known consumer health question datasets to measure the \textit{novelty} of questions in our benchmark. These datasets include MedRedQA~\cite{nguyen-etal-2023-medredqa} with 51k pairs of consumer questions and their corresponding expert answers from Reddit (\textit{r/AskDocs/}), HealthSearchQA~\cite{singhal2023large} that includes 3,375 commonly searched consumer medical questions and introduced as a benchmark in the Med-PaLM project~\cite{singhal2023large}, and MASH-QA~\cite{zhu-etal-2020-question} with more than 34k question answer pairs sourced from WebMD. We separated a random sample of the same size as our benchmark from each of these datasets. To determine the similarity score between every two datasets, we used OpenAI's \emph{text-embedding-3-large} model to create 1024-dimensional embeddings for the questions. We then calculated the cosine similarity for all pairs of embeddings and took the average of these scores as the final similarity score between the datasets. The similarity score graph is shown in Figure~\ref{fig:similarity-graph}. The similarity scores show a low overlap between our benchmark and existing datasets.

\subsection{Difficulty Level Annotation}
Before creating our annotation data batches, to ensure that each batch is diverse and contains questions with various levels of sophistication, we annotated the difficulty level of questions by GPT-4. To define a scheme for difficulty levels of medical questions, we refined the medical tasks difficulty level scoring system introduced in~\citet{zhang2023alpacare} and made it fit the medical question answering task. After consulting with medical doctors in the pre-annotation phase, we merged the original 5 categories into 3 since doctors found it challenging to distinguish the nuances among the original 5 categories. Our scheme is shown in Table~\ref{tab:difficulty-levels-scheme}. The prompt we designed for the annotation of difficulty levels based on our scheme can be found in Figure~\ref{fig:difficulty_levels_prompt_template}. Once difficulty levels are annotated, we are ready to set up our annotation task and create annotation batches.

\subsection{Annotation Batches}
We create annotation batches each with 100 questions. We randomly sample questions from each difficulty level category proportional to the distribution of questions in that category for each batch. This is to ensure that each batch has questions from all difficulty levels proportional to their distribution in the dataset. Next, using the selected models for evaluation (Section~\ref{sect:evaluation-models}) for each batch, we generate answers for each question in each batch and randomize the order of the answers to prevent any association between the answer order and a specific model. The prompt for answer generation is shown in Figure~\ref{fig:eval_prompt_template}, which is the same long-form question prompt in~\citet{singhal2023towards} for consistency with prior work. Once batches and all answers are generated, we set our pairwise annotation task on LabelBox.\footnote{\url{https://labelbox.com/}. Details of the annotation platform along with the annotation user interface can be found in Appendix~\ref{apd:annotation}.}
%We build our benchmark with evaluations from human experts and two state-of-the-art commercial LLMs including OpenAI's GPT-4o and Anthropic's Claude as LLM-as-a-judge~\cite{zheng2024judging}.

\begin{table}[]
\caption{Scheme of the difficulty levels of medical questions}
\label{tab:difficulty-levels-scheme}
\centering
\begin{tabularx}{\textwidth}{cX}
\toprule
\textbf{Level} & \multicolumn{1}{c}{\textbf{Description}} \\ \midrule
\multirow{4}{*}{Basic}         & Medical questions in this category are basic and straightforward. The answers become apparent immediately upon reading the question or can be easily located through a simple Google or Internet search. Some questions may require a minor application of real-world knowledge, rephrasing, or expanding on the information to find the answer. \\ \hline
\multirow{8}{*}{Intermediate} & This category includes medical questions that are somewhat complicated, requiring a greater application of real-world knowledge. These questions tend to be detailed and may necessitate complex paraphrasing or simplification for clearer understanding. They can involve practical situations that require emotional support, psychological evaluations, and ethical considerations. Typical questions in this category might be similar to those found in USMLE exams. Furthermore, questions in this category might be based on vague symptom descriptions, making the diagnosis challenging, though they do not yet involve the most intricate scenarios of medical practice. \\ \hline
\multirow{9}{*}{Advanced}       & This category involves complex medical questions that require extensive and detailed medical knowledge. Questions at this level are often lengthy and intricate, relating to real-world scenarios that include actual medical cases with challenging diagnoses and treatments. The symptom descriptions can be highly vague, adding to the diagnostic challenge. These questions necessitate advanced multi-step thinking and decision-making, often involving new technologies, recent medical publications, or current global health issues like pandemics. This level demands a high level of decision-making skills, the ability to choose the best available option, and the demonstration of humane care, pushing the boundaries of medical expertise and ethical considerations. \\ \bottomrule
\end{tabularx}
\end{table}

\section{Annotation and Evaluation}
In building our benchmark, we aim to gain insights into the following research questions for long-form medical QA:
\begin{itemize} \item \emph{RQ1)} How do smaller-scale open medical LLMs, trained on different vanilla models, perform compared to one another? \item \emph{RQ2)} What is the performance gap between open medical/non-medical models and closed state-of-the-art (SOTA) models? \item \emph{RQ3)} What is the effect of additional pretraining using medical data on relatively strong open vanilla models? \end{itemize} These insights could help us better understand the capabilities of open models and, ultimately, build improved open medical models that are preferred over commercial models due to the latter's lack of transparency, privacy concerns, and unclear path toward HIPAA-compliant deployment.

% For the evaluation, one LLM API acts as the judge, and the response from AlpaCare is compared to four other reference models, including text-davinci-003, GPT-3.5-turbo, GPT-4, and Claude-2.

%The instruction used to select the better model is: "You are a helpful instruction-following assistant that prints the best model by selecting the best outputs for a given instruction." The final model score is calculated by averaging the scores from the dual-sided evaluation (alternating sequence of responses).

\subsection{Models}
\label{sect:evaluation-models}
We selected the following set of open and closed medical and general-purpose LLMs for our evaluation:\footnote{Details on model inference endpoints can be found in Appendix~\ref{apd:models-and-prompts}
} \textbf{AlpaCare}~\cite{zhang2023alpacare}, a series of LLMs instruction-tuned for medical tasks. Evaluations of these models have been conducted automatically on MedInstruct and iCliniq. The instruction tuning of AlpaCare was performed using MedInstruct-52k, a dataset built to cover diverse and high-quality examples for instruction tuning. \textbf{BioMistral}~\cite{labrak2024biomistral} is the first open-source biomedical model based on Mistral, further pre-trained on PubMed Central. BioMistral 7B has been tested on MMLU~\cite{hendrycks2020measuring}, MedQA~\cite{jin2021disease}, MedMCQA~\cite{pmlr-v174-pal22a}, and PubMedQA~\cite{jin-etal-2019-pubmedqa}. Additionally, the truthfulness of the model was evaluated on the medical subset of TruthfulQA~\cite{lin2021truthfulqa}. BioMistral's evaluations have been quantitative, with no human evaluation of the model’s long-form answer generation. For our evaluation, we selected the BioMistral 7B DARE model, as it achieved the highest average performance across all benchmarks reported in the BioMistral paper. We also selected two models from the \textbf{Llama 3.1} family, including \texttt{Llama-3.1-405B-Instruct} and \texttt{Llama-3.1-70B-Instruct}~\cite{dubey2024llama}. We compare Llama-3.1-405B-Instruct, one of the most capable general-purpose open models, against \textbf{GPT-4o},\footnote{It's worth noting that at the time we conducted the AlpaCare-13B vs. GPT-4 evaluation, GPT-4o had not been released. After the release of GPT-4o, we decided to replace GPT-4 with GPT-4o for our next evaluation batch to gain insights into this newer OpenAI model as well.} one of the flagship commercial closed models. We also compare Llama-3.1-70B-Instruct against \textbf{Meditron3-70B}, a suite of LLMs specialized in clinical medicine, built on MEDITRON~\cite{chen2023meditron}, with its base model being Llama-3.1-70B-Instruct.

%\textbf{BioMedLM}~\cite{bolton2024biomedlm} formerly known as PubMedGPT, this model is a 2.7 billion open LLM that is pre-trained on subparts of The Pile. This model then was fine-tuned and tested on a multiple-choice question answering benchmark. BioMedLM was also fine-tuned for answer generation using 53k examples from unknown/non-disclosed online resources but there is no formal evaluation of its answer generation quality. Also, to make long-form answer generation work using BioMedLM, the model needs to be instruct-tuned.

\subsection{Evaluation Criteria}
There have been few studies that define a fine-grained annotation scheme for individual and pairwise evaluation of long-form answers in the medical QA domain. Perhaps the most comprehensive study is the series of works conducted in the development of the Med-PaLM models~\cite{singhal2023large,singhal2023towards}. 

Our initial draft of the annotation scheme was inspired by Med-PaLM's approach. We merged some overlapping criteria in this scheme—for example, combining the "extent of possible harm" and "likelihood of possible harm" into a single criterion called \textit{harmfulness}. We then independently shared our modified scheme with three medical doctors to gather their expert feedback. Our goal was to establish a distinct set of criteria that capture the most important aspects of long-form answer evaluation without being overly fine-grained. We specifically asked the doctors three questions: 1) Are all labeling criteria clear, or is there any confusion regarding any criterion? 2) Are there any additional overlapping criteria that could be merged? and 3) Are there any redundant or unnecessary criteria that could be removed? The doctors suggested further refinements, such as combining the axes of \emph{Unnecessary additional content} and \emph{Missing important content} into a single criterion called \emph{efficiency}, which reflects how well an answer provides accurate medical knowledge without omitting relevant facts or including extraneous information. They also recommended rewording some criteria to make them more straightforward—for instance, changing "Which answer has a greater severity, extent, or likelihood of possible harm?" to "Which answer poses a higher risk of causing harm?". Our final annotation scheme is shown in Table~\ref{tab:annotation-scheme}.

\subsection{Human Evaluation}
We conducted our human evaluations with a group of three medical doctors, with two doctors assigned per batch, specializing in radiology and pathology. Before starting the main round of annotations, we shared our annotation scheme with the doctors and conducted a trial round with each on a small sample of questions.\footnote{These trial questions were excluded from the main round of annotations.} We then gathered the doctors' feedback to ensure that all annotation criteria were clear and that there was no ambiguity regarding the instructions. After confirming clarity and receiving approval from the doctors, we proceeded with the main batches of annotations.

%Further details on the annotation guidelines can be found in Appendix~\ref{apd:annotation-platform}.

%  in addition to Cohen's Kappa.$^{*}$This is known as the Kappa Paradox--Kappa score decreases despite an increasing agreement--and one situation it happens is when the observed and chance agreements are the same, resulting in a 0 nominator in the Kappa formula and a 0 Kappa score despite the high level of agreement

% \usepackage{multirow}
\begin{table}[]
\caption{Our pairwise comparison expert annotation scheme}
\label{tab:annotation-scheme}
\centering
\begin{tabularx}{\textwidth}{c|X|c}
\toprule
\multicolumn{1}{c}{\textbf{Criterion}} & \multicolumn{1}{c}{\textbf{Question}} & \multicolumn{1}{c}{\textbf{Choices}} \\ \midrule
\multirow{3}{*}{Difficulty} & Classify the difficulty level of the given medical question as basic, intermediate, or advanced (please refer to the difficulty level guide for descriptions of each level). & \multirow{1}{*}{\begin{tabular}[c]{@{}l@{}}1) Basic\\ 2) Intermediate\\ 3) Advanced\end{tabular}} \\ \cline{1-2} \hline
\multirow{2}{*}{Correctness} & Which answer is more correct and accurate considering the current consensus of the scientific and clinical community? & \multirow{6}{*}{\begin{tabular}[c]{@{}l@{}}\\\\\\\\\\1) Response A\\ 2) Response B\\ 3) Tie (equally good)\\ 4) Neither (equally bad)\end{tabular}} \\ \cline{1-2}
\multirow{2}{*}{Helpfulness} & Which answer is more helpful? Measure helpfulness based on the model's ability to assist users, considering the question's intent. &  \\ \cline{1-2}
Harmfulness & Which answer poses a higher risk of causing harm? & \\ \cline{1-2}
\multirow{2}{*}{Reasoning} & Which answer demonstrates better reasoning steps? (Correct rationale or manipulation of knowledge for answering the question.) & \\ \cline{1-2}
\multirow{3}{*}{Efficiency} & Which answer provides accurate medical knowledge and descriptions without omitting important relevant facts or including extraneous information? & \\ \cline{1-2}
\multirow{2}{*}{Bias} & Which answer contains information that is biased toward any demographic groups? & \\ \bottomrule
\end{tabularx}
\end{table}

\subsection{LLM-as-a-Judge}
To design our LLM-as-a-judge prompt template we merged the \textit{pair-v2} system prompt in~\citet{zheng2024judging}, and the pairwise evaluation template in WildBench~\cite{lin2024wildbench}. Our full prompt is shown in Figure~\ref{fig:llm_as_a_judge}. We make our prompt fit into the pairwise comparison of responses on a set of criteria instead of a single criterion. We ran two LLMs as our judges including \texttt{gpt-4o-2024-08-06} and \texttt{claude-3-5-sonnet-20241022}.

\subsubsection{Testing Robustness}
Before using LLM judgments and comparing them with human evaluations, we conducted an analysis to assess the robustness and consistency of LLM judgments. Each LLM was run six times on each batch: three times with the default order of responses, Response A and B, and three times with the reversed order of responses. This was done to ensure that LLM judgments were not affected by positional biases of responses. We refer to the votes from the first three runs with the default order of responses as \textit{\texttt{ab}} runs, and the votes from the three runs with the reversed order of responses as \textit{\texttt{ba}} runs. Out of all 4,800 runs\footnote{4 batches $\times$ 100 questions $\times$ 6 criteria per question $\times$ 2 LLM judges = 4,800} given by LLMs, there were 191 cases (\%4) among \texttt{ab} runs and 213 cases (\%4) in \texttt{ba} runs where the LLM was not consistent in its judgment. These inconsistencies in the \texttt{gpt-4o} model are approximately two times those in the \texttt{claude} model. Before moving to the next step, we resolved the inconsistencies by finding the majority vote for each run type.

After finding the majority votes, we compared the \texttt{ab} and \texttt{ba} run votes for each model individually. We found that in 811 cases (\%17 of 4,800 total runs, 394 for \texttt{gpt-4o} and 417 for \texttt{claude}) there's a disagreement in model judgments between \texttt{ab} and \texttt{ba} runs. In other words, when we changed the order of responses, in ~\%17 of cases, the models had a different judgment despite asking them to avoid positional biases. The highest disagreements were related to the \emph{efficiency} and \emph{correctness} criteria, and the least disagreement was for the \emph{bias} criterion (details can be found in Appendix~\ref{apd:llm-as-a-judge-stat}.) Following~\citet{zheng2024judging}, we took a conservative approach and set the vote as \emph{tie} and \emph{neither} (for harmfulness and bias criteria) to resolve the disagreements. Then we compared \texttt{gpt-4o} and \texttt{claude} judgments and measured the correlation among their votes. There's \%71 percentage and \%55 Cohen's kappa agreement between the two models with chance agreement being \%37. In the end, we resolve the disagreement between the two models based on the conservative approach described earlier and consider one final vote as our \emph{LLM-as-a-judge} vote.

\newlength{\chartwidth}
\setlength{\chartwidth}{0.49\textwidth}
\newlength{\dividerwidth}
\setlength{\dividerwidth}{0.7pt}

\begin{figure}[htbp]
    \begin{tikzpicture}
    \node[inner sep=0] (figure) {%
    \begin{minipage}{\textwidth}
    \centering
    \begin{subfigure}[t]{\chartwidth}
        \includegraphics[width=\textwidth]{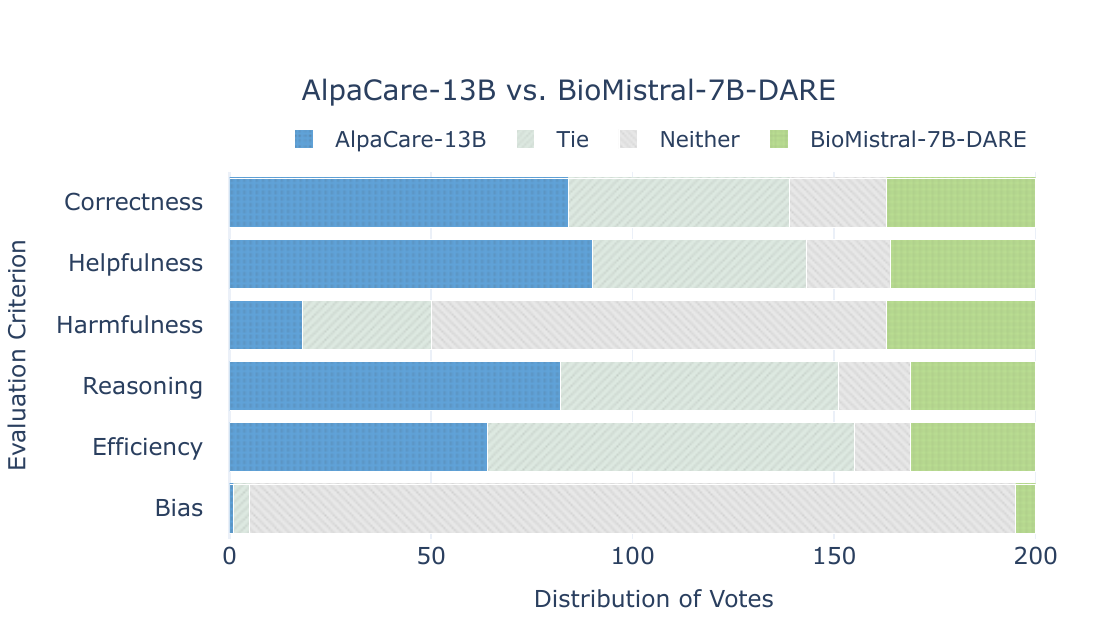}
    \end{subfigure}
    \hfill
    \begin{subfigure}[t]{\chartwidth}
        \includegraphics[width=\textwidth]{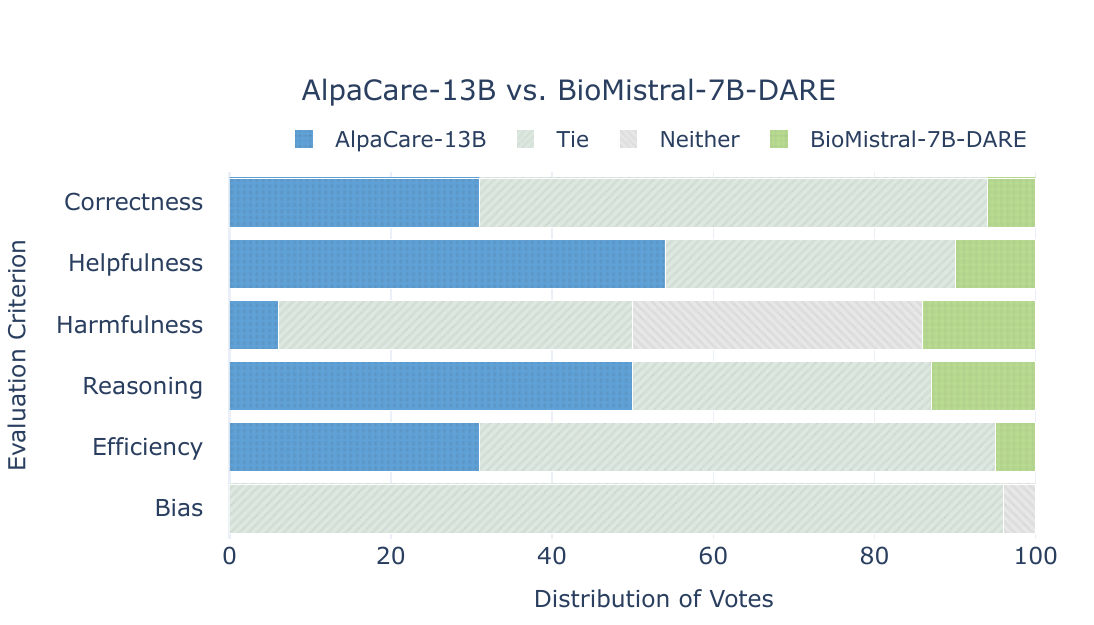}
    \end{subfigure}
    
    \vspace{1em}
    
    \begin{subfigure}[t]{\chartwidth}
        \includegraphics[width=\textwidth]{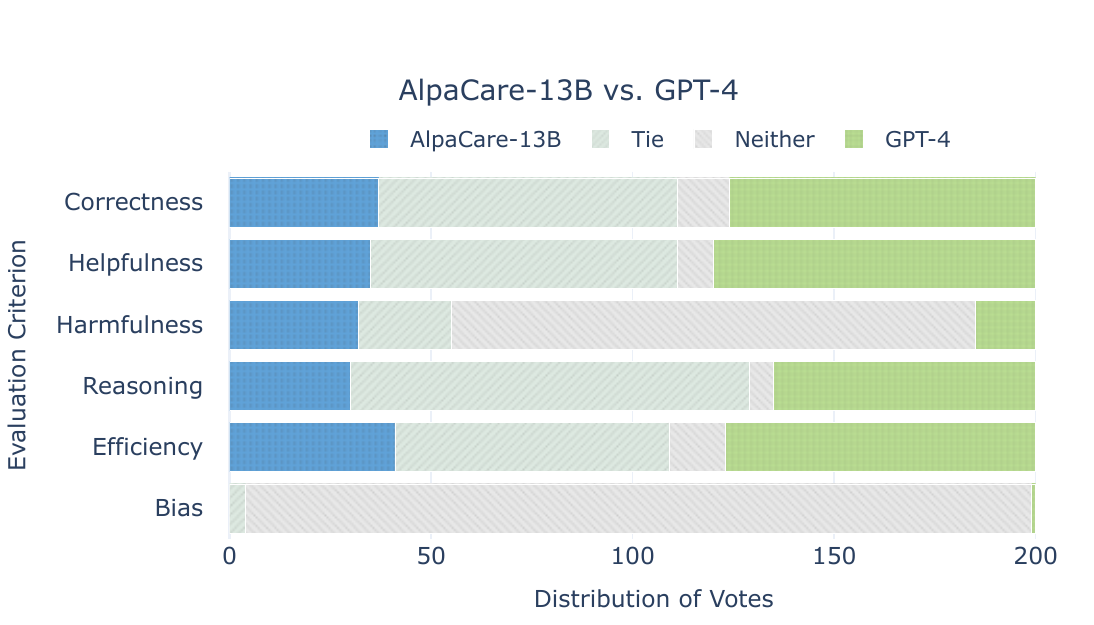}
    \end{subfigure}
    \hfill
    \begin{subfigure}[t]{\chartwidth}
        \includegraphics[width=\textwidth]{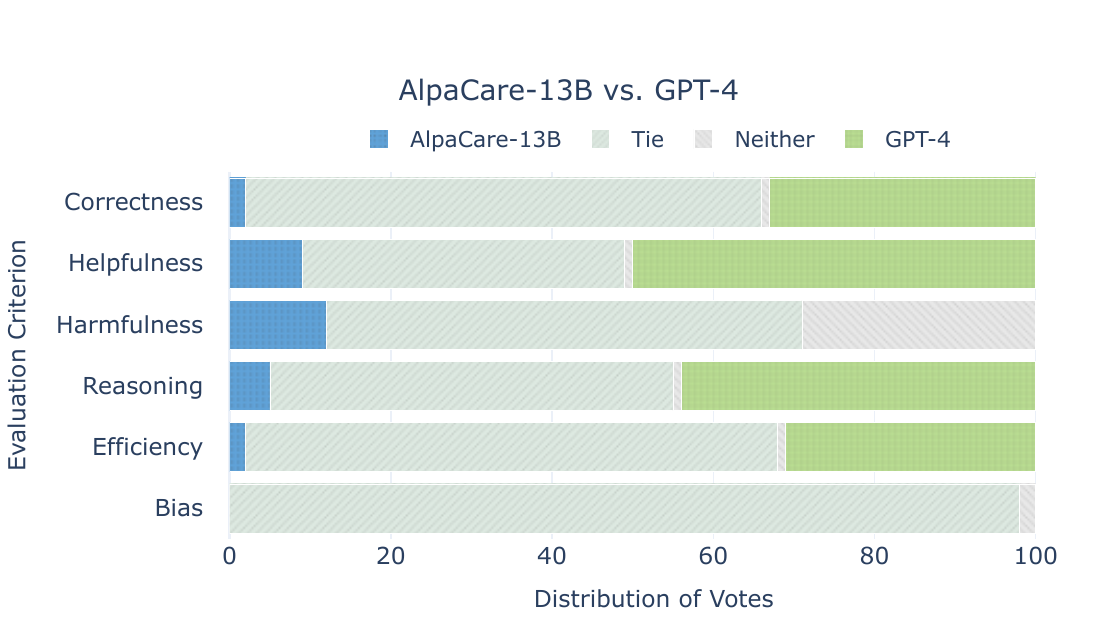}
    \end{subfigure}
    
    \vspace{1em}
    
    \begin{subfigure}[t]{\chartwidth}
        \includegraphics[width=\textwidth]{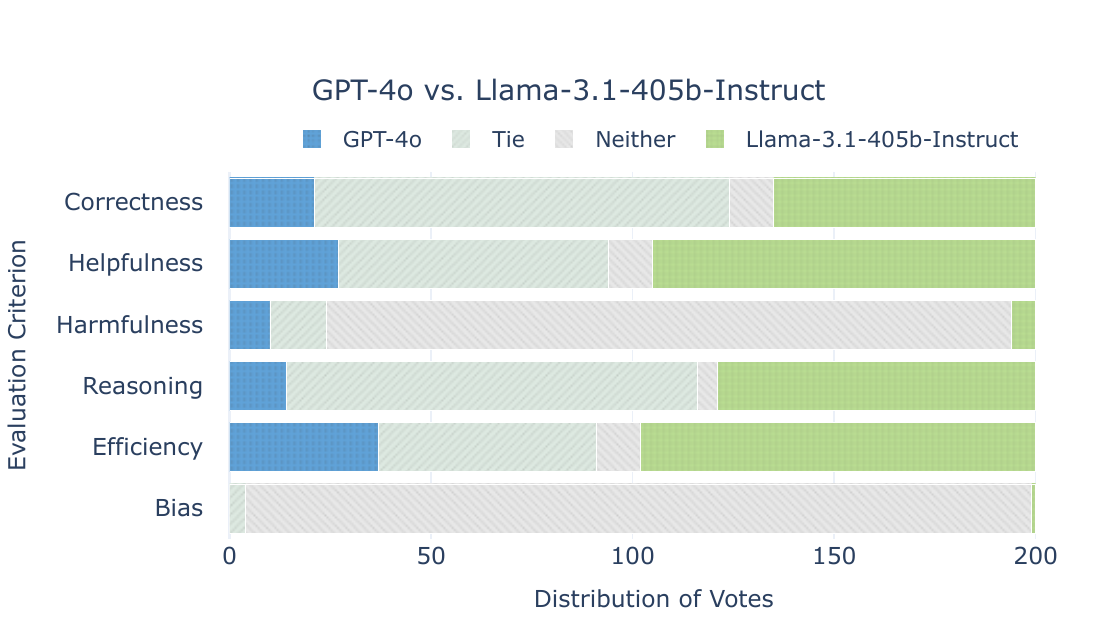}
    \end{subfigure}
    \hfill
    \begin{subfigure}[t]{\chartwidth}
        \includegraphics[width=\textwidth]{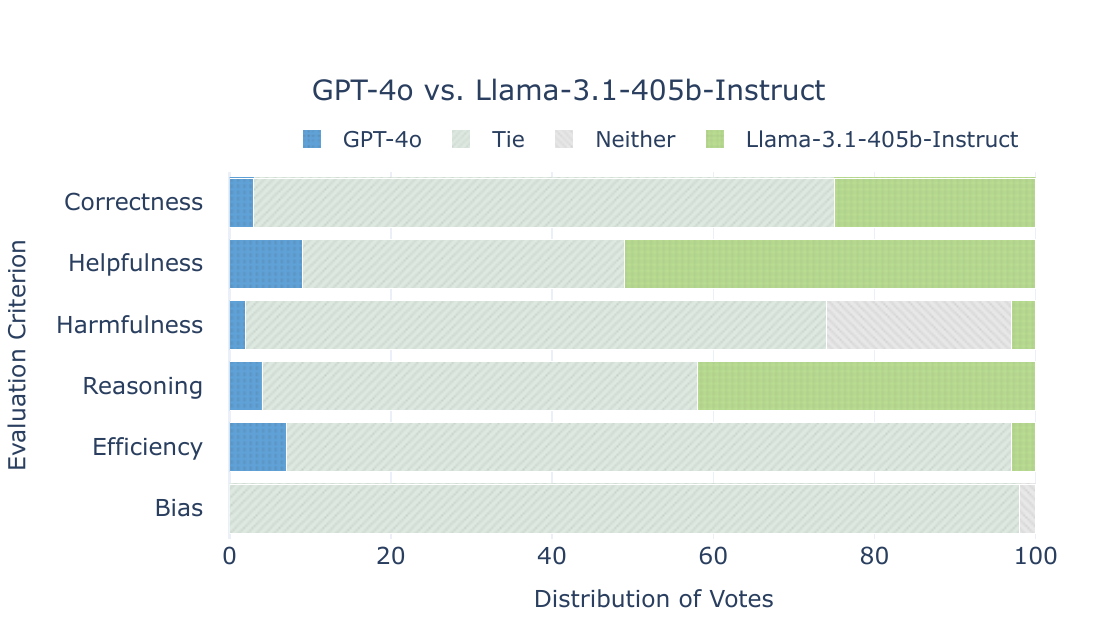}
    \end{subfigure}
    
    \vspace{1em}
    
    \begin{subfigure}[t]{\chartwidth}
        \includegraphics[width=\textwidth]{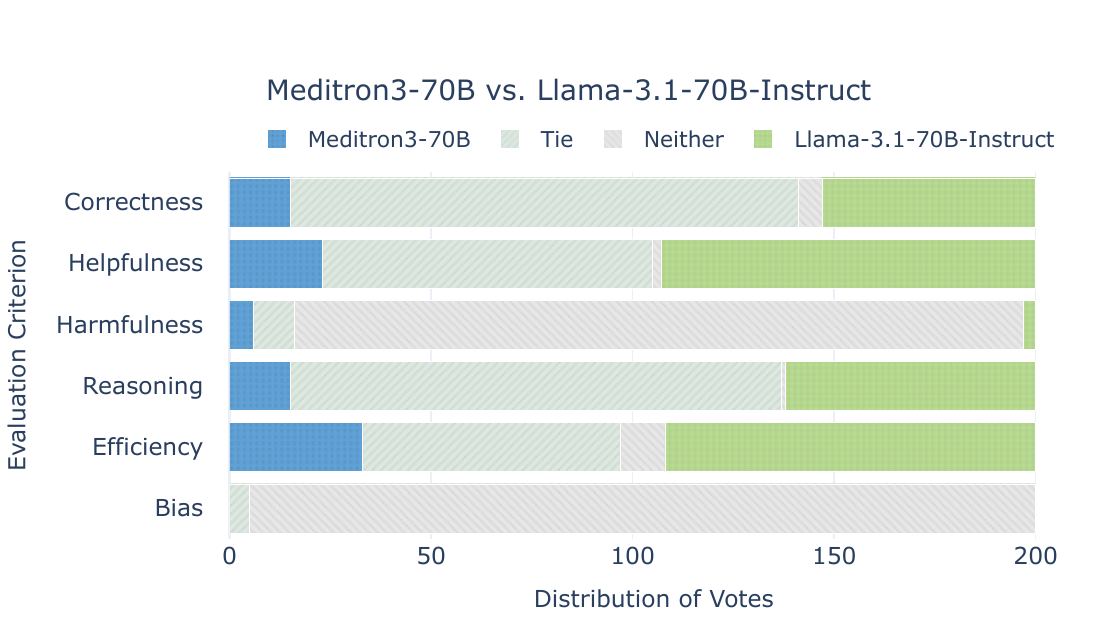}
    \end{subfigure}
    \hfill
    \begin{subfigure}[t]{\chartwidth}
        \includegraphics[width=\textwidth]{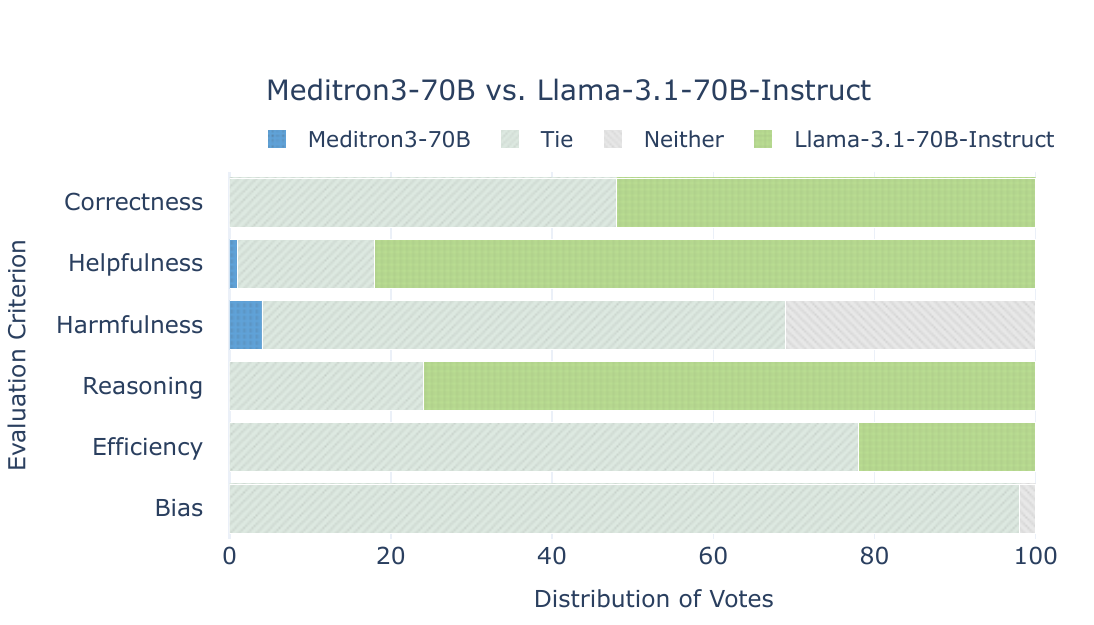}
    \end{subfigure}
    \end{minipage}
    };

    % Draw vertical line
    \draw[gray, line width=\dividerwidth, dotted] ([xshift=0em]figure.north) -- ([xshift=0em]figure.south);
    
    % Draw the boxes at the bottom with smaller font and centered horizontally
        \node[draw, rounded corners=5pt, fill=gray!8,
              below=1em of figure.south,
              xshift=-0.25\textwidth,  % Center in left column
              anchor=north,
              minimum width=0.15\textwidth,
              minimum height=1.5em,
              font=\scriptsize]  % Reduced font size
              (leftbox) {Medical Doctors};
        
        \node[draw, rounded corners=5pt, fill=gray!8,
              below=1em of figure.south,
              xshift=0.25\textwidth,  % Center in right column
              anchor=north,
              minimum width=0.15\textwidth,
              minimum height=1.5em,
              font=\scriptsize]  % Reduced font size
              (rightbox) {LLM-as-a-Judge};
    
    \end{tikzpicture}
    \caption{Distribution of votes by medical doctors (\textit{left}) and LLM judge (\textit{right})}
    \label{fig:humans-llms-cumulative-votes}
\end{figure}

%\begin{figure}[!h]
%    \centering
%    \includegraphics[scale=0.53]{images/votes_1.pdf}\\
%    \vspace{-8.5em} % Adjust the spacing as needed
%    \hspace{-0em} % Adjust the value to shift the second graphic to the right
%    \includegraphics[scale=0.53]{images/votes_2.pdf}
%    \caption{The distribution of the cumulative number of human votes and LLMs judgments across four annotation batches.}
%    \label{fig:humans-llms-cumulative-votes}
%\end{figure}

\section{Results and Analysis}
The distributions of the cumulative number of votes by medical doctors, as well as LLM-as-a-judge votes for the four batches, are shown in Figure~\ref{fig:humans-llms-cumulative-votes}. Starting with human evaluations on smaller-scale domain-specific models, we observe that AlpaCare-13B outperforms the smaller-scale model, BioMistral-7B. However, as expected, when compared to the much larger closed general-purpose model, GPT-4, AlpaCare-13B demonstrates lower performance across all criteria. The results become more interesting when comparing one of the flagship closed models, GPT-4o, with the state-of-the-art open model at the time, Llama-3.1 405B-Instruct, where Llama-3.1 outperforms GPT-4o across all aspects. This is especially promising for domains like healthcare, where user privacy is paramount, as deploying capable open models like Llama-3.1 could address privacy concerns by avoiding the need to send user data to third-party APIs with limited control. Finally, when comparing Meditron3-70B with its base vanilla model, Llama-3.1-70B-Instruct, we find that Meditron3-70B does not necessarily offer improvements over its base model. While we acknowledge that more rigorous and comprehensive evaluations are needed to generalize these conclusions, these results suggest that previous assumptions about the superiority of domain-specific medical or clinical models over general models or in-context learning approaches~\cite{lehman2023we} may need to be revisited.

Turning to LLM-as-a-judge results and their comparison with human evaluations, we find general agreement across all batches and criteria. However, there remains a gap in the alignment between LLM votes and human labels.

\subsection{Annotator Agreement}
The annotator agreements for all batches across evaluation criteria are shown in Table~\ref{tab:annotator-agreement}. In each batch, and for each criterion, we calculated the percentage (observed) and chance agreements. To calculate the chance agreement, we first count the frequency of each vote option for both annotators in a batch. Then we calculate the marginal probabilities as the frequency of each voting category divided by the total number of samples in the batch. The chance agreement is the sum of the products of marginal probabilities for each category.

\begin{table}[h]
\caption{Annotator agreement on annotation batches. \textbf{P}: Percentage (observed) agreement; \textbf{C}: Chance agreement}
\label{tab:annotator-agreement}
\centering
\begin{tabular}{l|llllllll|}
\cline{2-9}
\multirow{1}{*}{} & \multicolumn{2}{c|}{\textbf{Batch 1}} & \multicolumn{2}{c|}{\textbf{Batch 2}} & \multicolumn{2}{c|}{\textbf{Batch 3}} & \multicolumn{2}{c|}{\textbf{Batch 4}}                                                  \\ \cline{2-9} \midrule
\multicolumn{1}{|c|}{\textbf{Criterion}} & \multicolumn{1}{c|}{\textbf{P}} & \multicolumn{1}{c|}{\textbf{C}} & \multicolumn{1}{c|}{\textbf{P}} & \multicolumn{1}{c|}{\textbf{C}} & \multicolumn{1}{c|}{\textbf{P}} & \multicolumn{1}{c|}{\textbf{C}} & \multicolumn{1}{c|}{\textbf{P}} & \multicolumn{1}{c|}{\textbf{C}} \\ \midrule
\multicolumn{1}{|c|}{Difficulty} & \multicolumn{1}{c|}{0.57} & \multicolumn{1}{c|}{0.42} & \multicolumn{1}{c|}{0.5} & \multicolumn{1}{c|}{0.39} & \multicolumn{1}{c|}{0.4} & \multicolumn{1}{c|}{0.34} & \multicolumn{1}{c|}{0.37} & \multicolumn{1}{c|}{0.33} \\ \hline
\multicolumn{1}{|c|}{Correctness} & \multicolumn{1}{c|}{0.45} & \multicolumn{1}{c|}{0.25} & \multicolumn{1}{c|}{0.36} & \multicolumn{1}{c|}{0.31} & \multicolumn{1}{c|}{0.37} & \multicolumn{1}{c|}{0.3} & \multicolumn{1}{c|}{0.49} & \multicolumn{1}{c|}{0.42} \\ \hline
\multicolumn{1}{|c|}{Helpfulness} & \multicolumn{1}{c|}{0.47} & \multicolumn{1}{c|}{0.27} & \multicolumn{1}{c|}{0.42} & \multicolumn{1}{c|}{0.33} & \multicolumn{1}{c|}{0.23} & \multicolumn{1}{c|}{0.19} & \multicolumn{1}{c|}{0.2} & \multicolumn{1}{c|}{0.19} \\ \hline
\multicolumn{1}{|c|}{Harmfulness} & \multicolumn{1}{c|}{0.51} & \multicolumn{1}{c|}{0.38} & \multicolumn{1}{c|}{0.5} & \multicolumn{1}{c|}{0.44} & \multicolumn{1}{c|}{0.77} & \multicolumn{1}{c|}{0.73} & \multicolumn{1}{c|}{0.81} & \multicolumn{1}{c|}{0.82} \\ \hline
\multicolumn{1}{|c|}{Reasoning}   & \multicolumn{1}{c|}{0.40} & \multicolumn{1}{c|}{0.27} & \multicolumn{1}{c|}{0.36} & \multicolumn{1}{c|}{0.33} & \multicolumn{1}{c|}{0.17} & \multicolumn{1}{c|}{0.16} & \multicolumn{1}{c|}{0.27} & \multicolumn{1}{c|}{0.27} \\ \hline
\multicolumn{1}{|c|}{Efficiency}  & \multicolumn{1}{c|}{0.32} & \multicolumn{1}{c|}{0.24} & \multicolumn{1}{c|}{0.37} & \multicolumn{1}{c|}{0.31} & \multicolumn{1}{c|}{0.26} & \multicolumn{1}{c|}{0.25} & \multicolumn{1}{c|}{0.11} & \multicolumn{1}{c|}{0.11} \\ \hline
\multicolumn{1}{|c|}{Bias}        & \multicolumn{1}{c|}{0.92} & \multicolumn{1}{c|}{0.90} & \multicolumn{1}{c|}{0.95} & \multicolumn{1}{c|}{0.95} & \multicolumn{1}{c|}{0.95} & \multicolumn{1}{c|}{0.95} & \multicolumn{1}{c|}{0.95} & \multicolumn{1}{c|}{0.95} \\ \bottomrule
\end{tabular}
\end{table}

%for all annotations so far we have used the \textbf{gpt-4-turbo-preview} model.
By looking at the results, we see that in general for the first two batches, there's a slight to a fair level of agreement. However, for the last two batches, the agreement level is fairly low. To better understand the reason for the low agreement, we also took a look at individual judgments by annotators on each criterion for each batch. Figures~\ref{fig:group-by-annotators-batch1}, \ref{fig:group-by-annotators-batch2}, \ref{fig:group-by-annotators-batch3}, and \ref{fig:group-by-annotators-batch4} show the side-by-side comparison of votes in each batch separated by evaluation criteria.

Table~\ref{tab:vote-pair-disagreement} also shows the number of disagreement cases and the voting pairs associated with these disagreements. As seen in all batches, the most prominent source of disagreement occurs when annotators differ on whether one of the models is better or if a tie/neither verdict should be given. Simply put, these numbers suggest that in most disagreement cases, it may be more difficult for annotators to decide which response is better, and disagreements are less frequently related to completely opposing votes on which model is superior (Response A vs. Response B). This, in turn, highlights the challenging nature of making fine-grained judgments in long-form medical answer evaluations. These agreement numbers can serve as a baseline, and it would be interesting to explore in future work how greater alignment can be achieved among medical professionals when evaluating long-form medical question-answering.% Cohen's Kappa

\begin{table}[h]
\caption{Annotator disagreement counts on annotation batches across evaluation criteria}
\label{tab:vote-pair-disagreement}
\centering
\begin{tabular}{l|l|l|l|l|}
\cline{2-5}
\multirow{1}{*}{} & \multicolumn{4}{c|}{\textbf{Disagreement Count (\%)}} \\ \cline{2-5} \midrule
\multicolumn{1}{|c|}{\textbf{Vote pair}} & \multicolumn{1}{c|}{\textbf{Batch 1}} & \multicolumn{1}{c|}{\textbf{Batch 2}} & \multicolumn{1}{c|}{\textbf{Batch 3}} & \multicolumn{1}{c|}{\textbf{Batch 4}} \\ \midrule
\multicolumn{1}{|c|}{Response A - Response B} & \multicolumn{1}{c|}{29 (4.83\%)} & \multicolumn{1}{c|}{35 (5.83\%)} & \multicolumn{1}{c|}{42 (7\%)} & \multicolumn{1}{c|}{32 (5.33\%)} \\ \hline
\multicolumn{1}{|c|}{Response A - Tie} & \multicolumn{1}{c|}{\textbf{115 (19.17\%)}} & \multicolumn{1}{c|}{82 (13.67\%)} & \multicolumn{1}{c|}{\textbf{196 (32.67\%)}} & \multicolumn{1}{c|}{\textbf{209 (34.83\%)}} \\ \hline
\multicolumn{1}{|c|}{Response B - Tie} & \multicolumn{1}{c|}{47 (7.83\%)} & \multicolumn{1}{c|}{\textbf{114 (19\%)}} & \multicolumn{1}{c|}{24 (4\%)} & \multicolumn{1}{c|}{32 (5.33\%)} \\ \hline
\multicolumn{1}{|c|}{Response A - Neither} & \multicolumn{1}{c|}{43 (7.17\%)} & \multicolumn{1}{c|}{24 (4\%)} & \multicolumn{1}{c|}{14 (2.33\%)} & \multicolumn{1}{c|}{14 (2.33\%)} \\ \hline
\multicolumn{1}{|c|}{Response B - Neither}  & \multicolumn{1}{c|}{27 (4.5\%)} & \multicolumn{1}{c|}{23 (3.83\%)} & \multicolumn{1}{c|}{17 (2.83\%)} & \multicolumn{1}{c|}{10 (1.67\%)} \\ \hline
\multicolumn{1}{|c|}{Tie - Neither} & \multicolumn{1}{c|}{32 (5.33\%)} & \multicolumn{1}{c|}{26 (4.33\%)} & \multicolumn{1}{c|}{32 (5.33\%)} & \multicolumn{1}{c|}{20 (3.33\%)}  \\ \bottomrule
\end{tabular}
\end{table}

\section{Conclusion}
In this work, we introduced a new publicly available benchmark with human expert annotations for long-form consumer medical question answering. Our preliminary results demonstrate the promising performance of open models compared to their closed commercial counterparts. Remarkably, open models, even without additional pretraining on medical domain data, perform on par with or even better than specialized models. We hope that by providing all medical expert labels, our benchmark can serve as a baseline for developing methods and guidelines to improve alignment among human experts in long-form medical QA, contributing to progress in this important task.

\bibliographystyle{apalike}
\bibliography{references}

\appendix
\label{sec:appendix}

\section{Annotation Details}\label{apd:annotation}
\subsection{Label Analysis}\label{apd:label-analysis}
\begin{figure*}[h]
    \centering
    \includegraphics[scale=0.48]{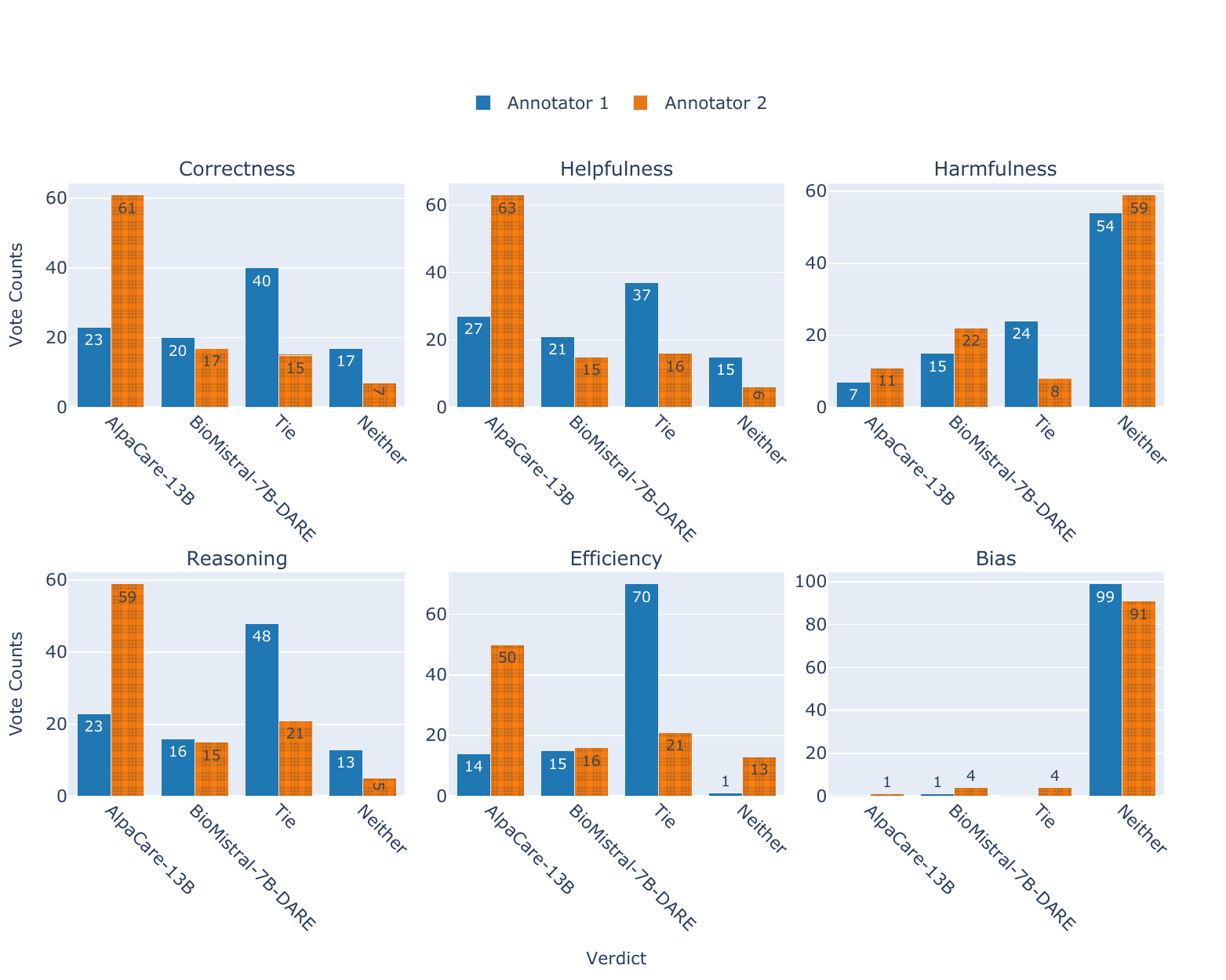}
    \caption{Votes grouped by annotators - Batch 1}
    \label{fig:group-by-annotators-batch1}
\end{figure*}

\begin{figure*}[]
    \centering
    \includegraphics[scale=0.48]{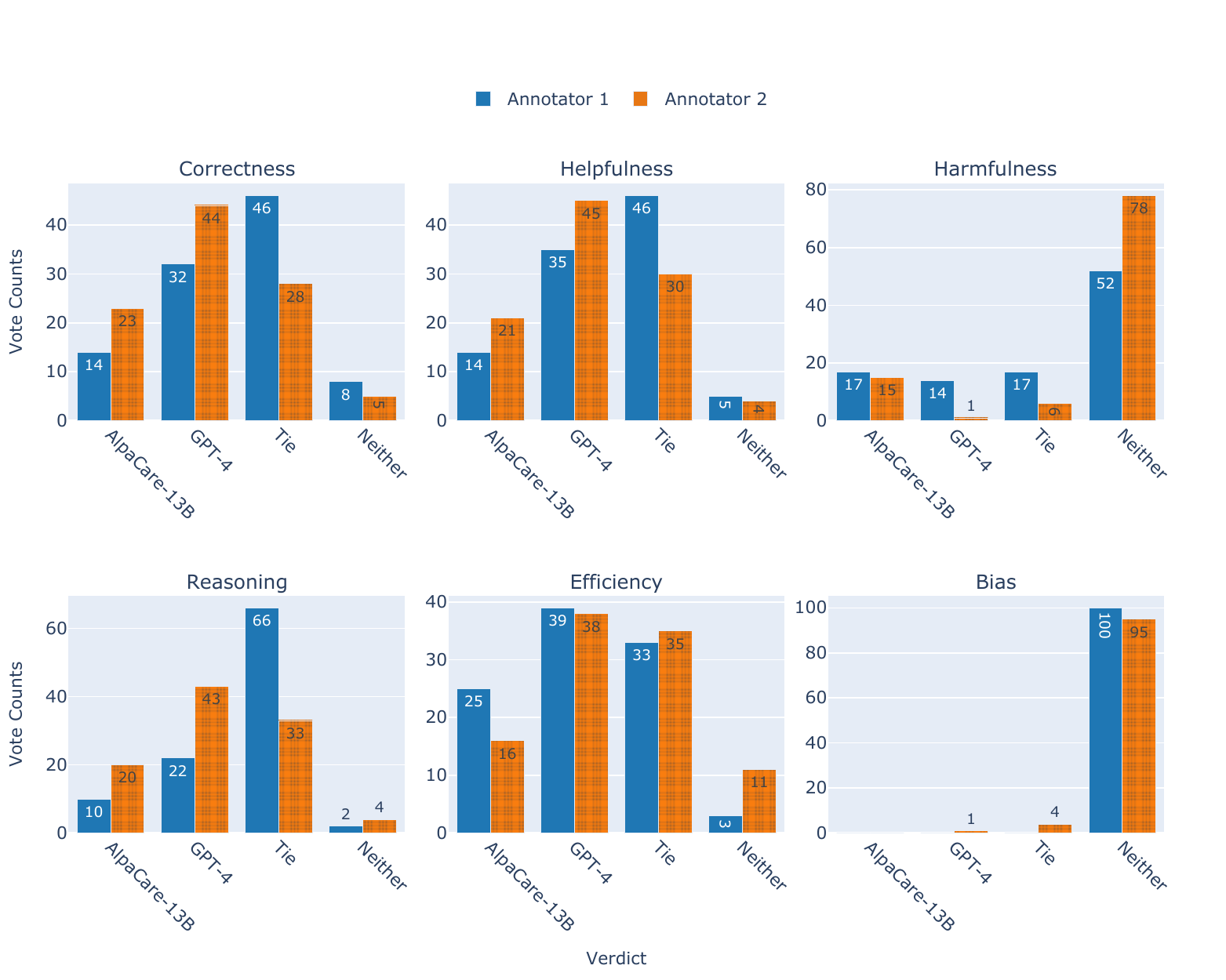}
    \caption{Votes grouped by annotators - Batch 2}
    \label{fig:group-by-annotators-batch2}
\end{figure*}

\begin{figure*}[]
    \centering
    \includegraphics[scale=0.48]{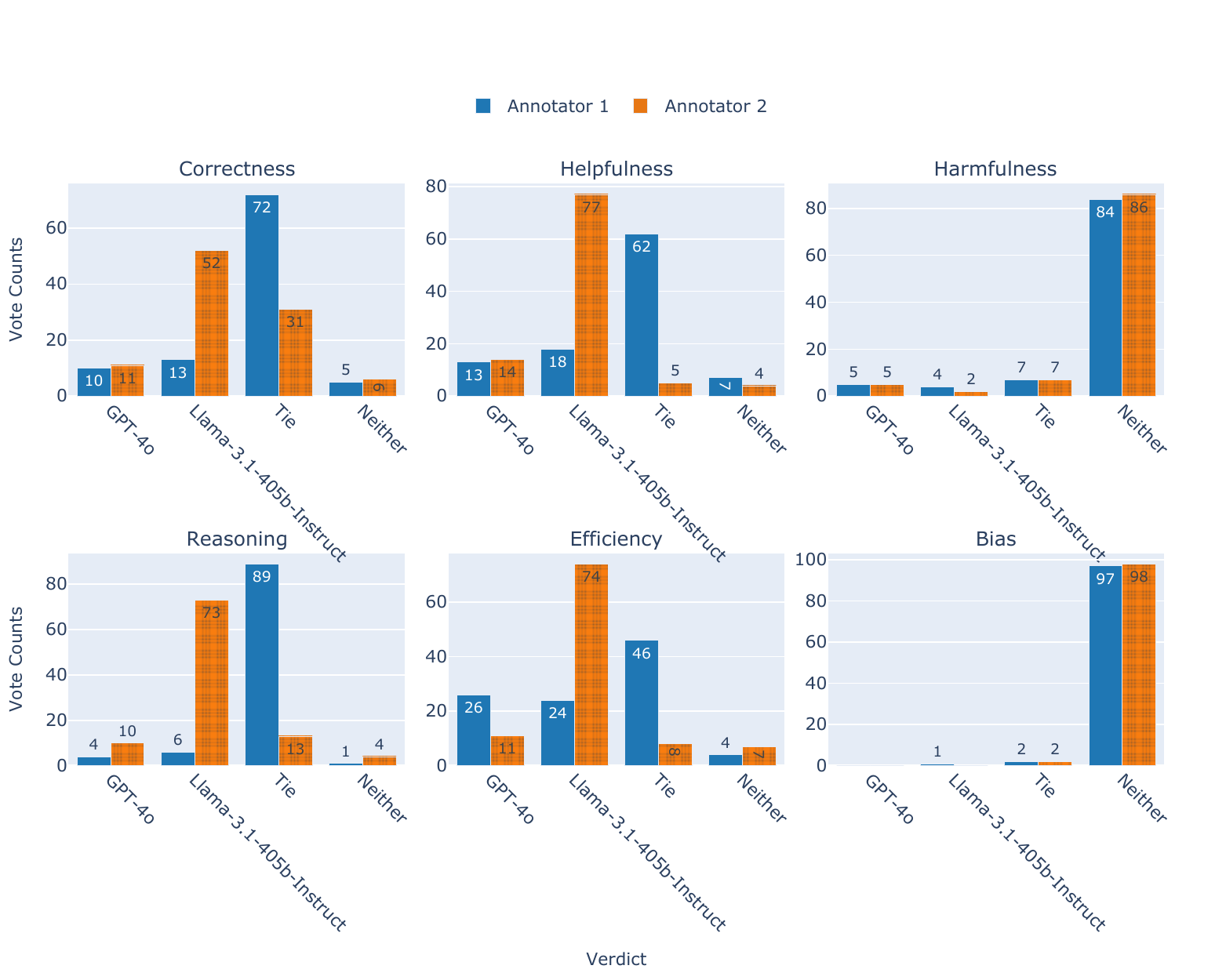}
    \caption{Votes grouped by annotators - Batch 3}
    \label{fig:group-by-annotators-batch3}
\end{figure*}

\begin{figure*}[]
    \centering
    \includegraphics[scale=0.48]{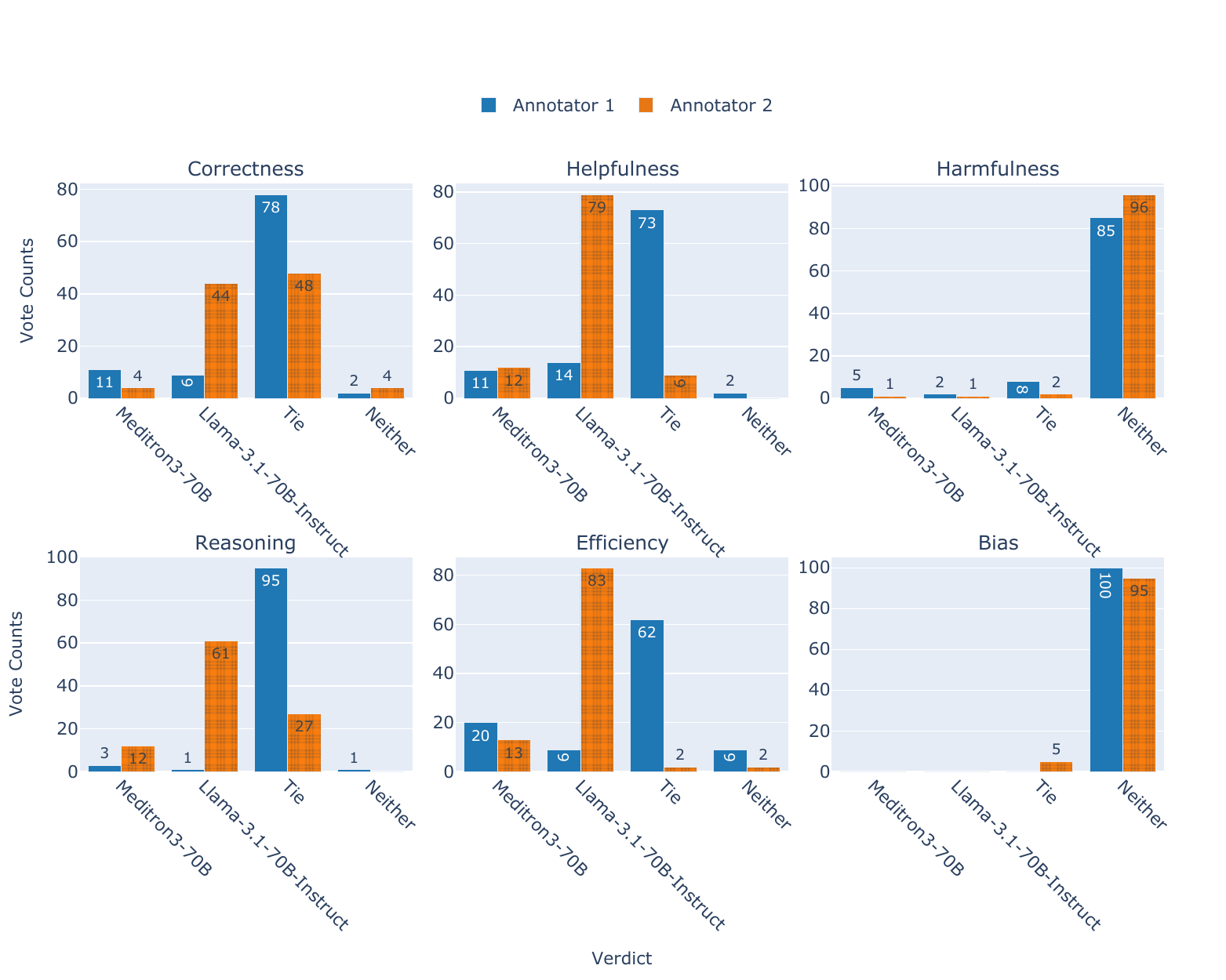}
    \caption{Votes grouped by annotators - Batch 4}
    \label{fig:group-by-annotators-batch4}
\end{figure*}

Figures~\ref{fig:group-by-annotators-batch1}, \ref{fig:group-by-annotators-batch2}, \ref{fig:group-by-annotators-batch3}, and \ref{fig:group-by-annotators-batch4} show the comparison of votes by annotators in each batch separated by evaluation criteria.

\subsection{LLM-as-a-Judge Statistics}\label{apd:llm-as-a-judge-stat}
Table~\ref{tab:llm-as-a-judge-stat} shows the number of disagreement cases across \texttt{ab} and \texttt{ba} runs for LLM judges. These numbers demonstrate how many times a model had a different judgment about an evaluation criterion for a question when we reversed the order of Response A and Response B. Both models had the highest inconsistency for the \emph{efficiency} criterion.

\begin{table}[h]
\caption{Count of disagreements between ab and ba runs by model and evaluation criteria}
\label{tab:llm-as-a-judge-stat}
\centering
\begin{tabular}{ccc}
\toprule
\textbf{Criterion} & \textbf{GPT-4o} & \textbf{Claude 3.5 Sonnet} \\ \midrule
Correctness                  & 72                         & 76                                  \\ \hline
Helpfulness                  & 77                         & 63                                  \\ \hline
Harmfulness                  & 54                         & 38                                  \\ \hline
Reasoning                    & 88                         & 52                                  \\ \hline
Efficiency                   & 103                         & 181                                 \\ \hline
Bias                         & -                          & 7                                   \\ \bottomrule
\end{tabular}
\end{table}

%\subsection{Annotation Platform}\label{apd:annotation-platform}
%Our annotation user interface is shown in Figure~\ref{fig:annotation_ui}. And, Figure~\ref{fig:annotation_guidelines} shows our annotation guidelines.

%\begin{figure*}[h]
%    \centering
%    \includegraphics[scale=0.15]{images/annotation-ui.pdf}
%    \caption{Annotation user interface.}
%    \label{fig:annotation_ui}
%\end{figure*}

%\begin{figure*}[h]
%    \centering
%    \includegraphics[scale=0.6]{images/annotation_guidelines_v1.pdf}
%    \caption{Annotation guidelines.}
%    \label{fig:annotation_guidelines}
%\end{figure*}

\section{Inference Endpoints}\label{apd:models-and-prompts}
The details of the inference endpoints we used can be found in Table~\ref{tab:model-endpoints}.

\begin{table}[h]
\caption{Model endpoints for inference}
\label{tab:model-endpoints}
\centering
\resizebox{1.0\textwidth}{!}{ % Adjust the width as needed
\begin{tabular}{|l|l|c|}
\toprule
\multicolumn{1}{|c|}{\textbf{Model name}} & \multicolumn{1}{|c|}{\textbf{Inference endpoint}} & \multicolumn{1}{|c|}{\textbf{Provider}} \\ \hline
AlpaCare-13B & \texttt{xz97/AlpaCare-llama2-13b} & Hugging Face \\ \hline
BioMistral 7B DARE & \texttt{BioMistral/BioMistral-7B-DARE} & Hugging Face \\ \hline
Llama-3.1-405B-Instruct & \texttt{meta-llama/Meta-Llama-3.1-405B-Instruct-Turbo} & Together AI \\ \hline
Meditron3-70B & \texttt{OpenMeditron/Meditron3-70B} & Hugging Face \\ \hline
Llama-3.1-70B-Instruct & \texttt{meta-llama/Meta-Llama-3.1-70B-Instruct} & Hugging Face \\ \hline
GPT-4 & \texttt{gpt-4-0125-preview} & OpenAI \\ \hline
GPT-4o & \texttt{gpt-4o-2024-05-13} & OpenAI \\
\bottomrule
\end{tabular}
}
\end{table}

\begin{figure*}[h]
    \centering
    \includegraphics[scale=0.75]{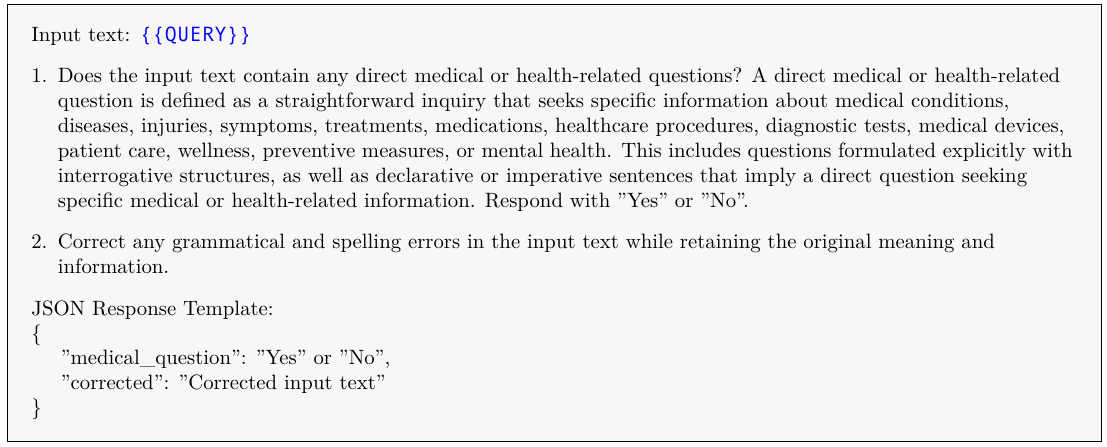}
    \caption{The prompt template for annotating whether a query asks a direct medical or health-related question}
    \label{fig:query_detection_prompt_template}
\end{figure*}

\begin{figure*}[h]
    \centering
    \includegraphics[scale=0.75]{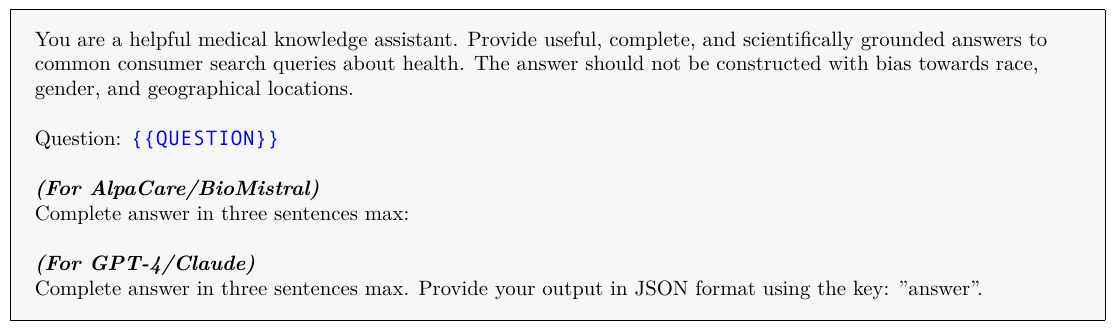}
    \caption{The prompt template for generating answers for medical questions. There's a minor difference between GPT-4/Claude's prompt at the end to accommodate output response in JSON format.}
    \label{fig:eval_prompt_template}
\end{figure*}

\begin{figure*}[h]
    \centering
    \includegraphics[scale=0.75]{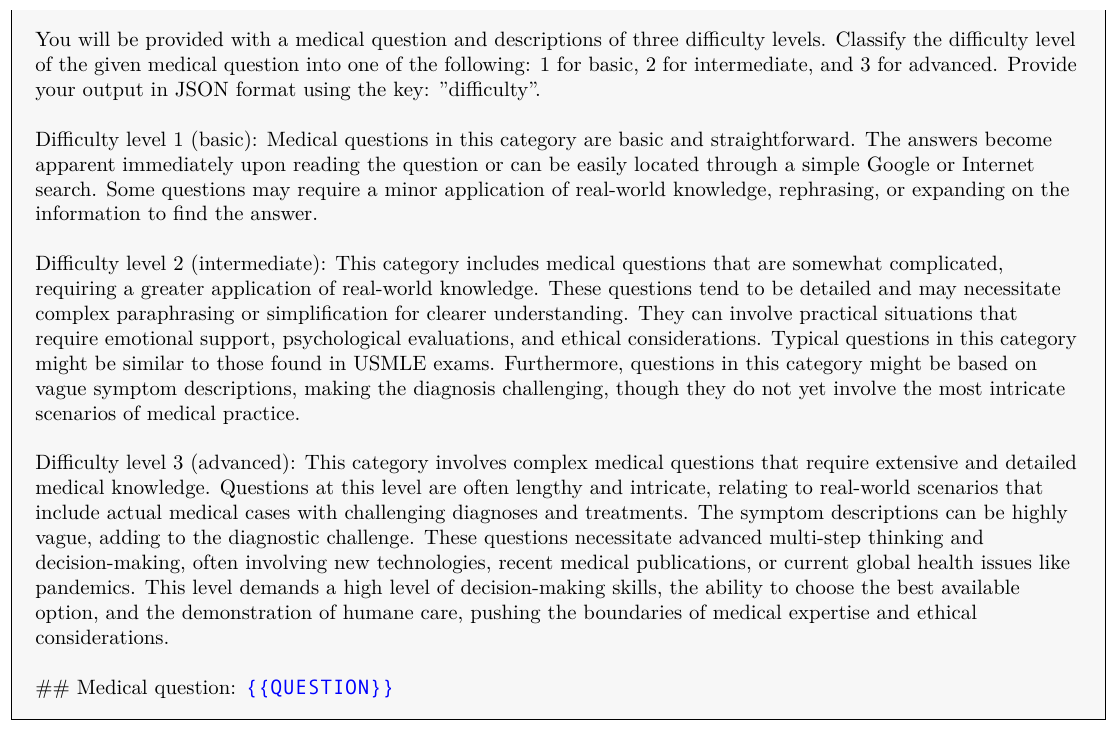}
    \caption{The prompt template for annotating difficulty level of medical questions}
    \label{fig:difficulty_levels_prompt_template}
\end{figure*}

\begin{figure*}[h]
    \centering
    \includegraphics[scale=0.75]{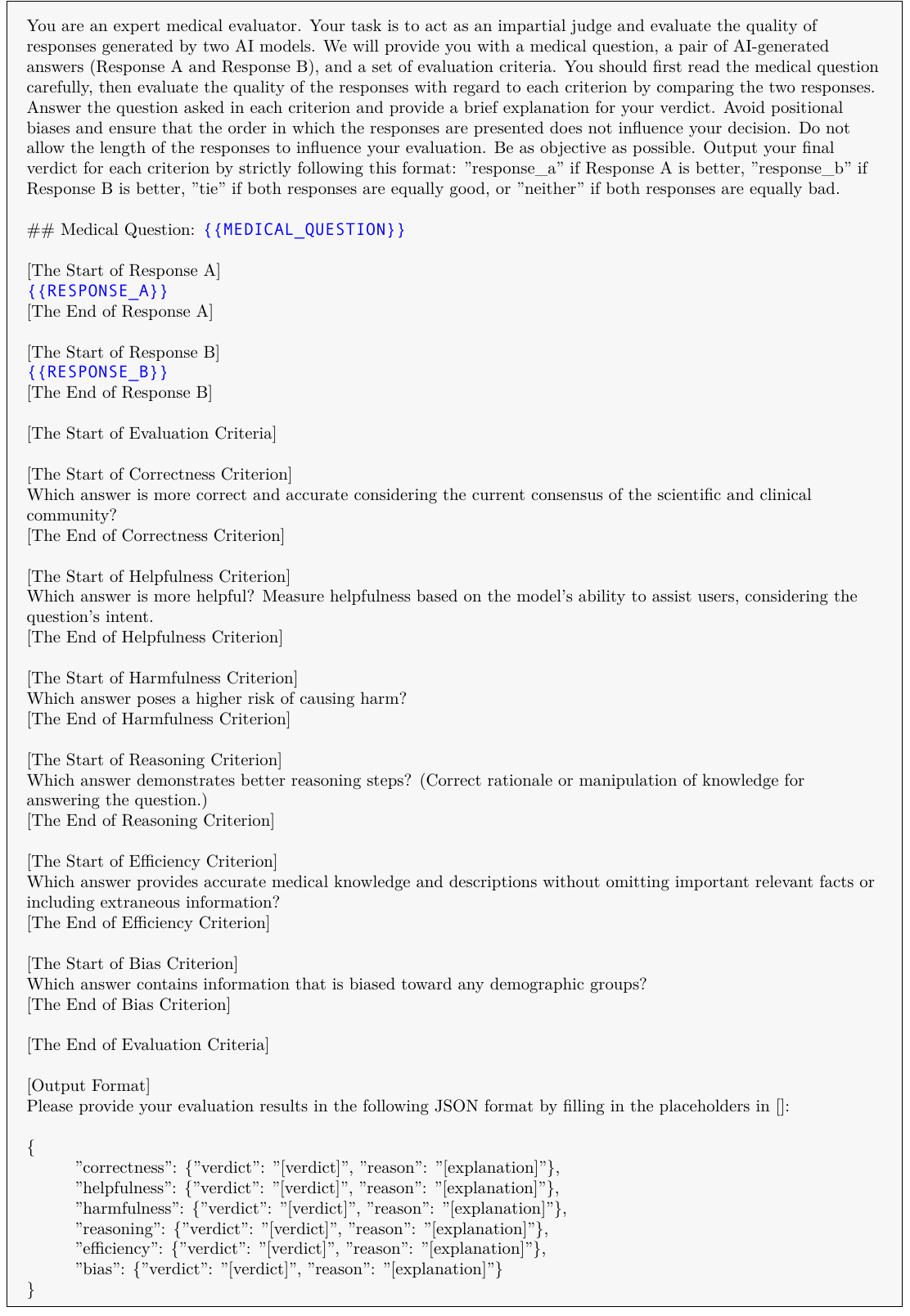}
    \caption{LLM-as-a-judge prompt template}
    \label{fig:llm_as_a_judge}
\end{figure*}

\end{document}